\pdfoutput=1

\documentclass[11pt]{article}

\usepackage[]{EMNLP2022}

\usepackage{times}
\usepackage{latexsym}

\usepackage[T1]{fontenc}

\usepackage[utf8]{inputenc}

\usepackage{microtype}

\usepackage{inconsolata}

\usepackage{amsmath}
\usepackage{mathrsfs,amssymb}
\usepackage{enumitem}
\usepackage{color}
\usepackage{graphicx,booktabs}
\graphicspath{figures}
\usepackage{caption} 
\usepackage[labelformat=simple]{subcaption}

\usepackage{xspace}
\usepackage{array}
\usepackage{xurl}
\usepackage{multirow}
\usepackage{todonotes}

\usepackage{adjustbox}

\title{
Investigating Ensemble Methods for Model Robustness Improvement \\ of Text Classifiers}

\author{Jieyu Zhao$^{1}$\thanks{\hspace{1.5mm}Work was done while interning at Google Research.} \quad Xuezhi Wang$^{2}$ \quad Yao Qin$^{2}$ \quad Jilin Chen$^{2}$ \quad Kai-Wei Chang$^{1}$
\\
  $^{1}$University of California, Los Angeles  \qquad
  $^{2}$Google Research \\
  jieyuzhao@ucla.edu  \quad  \{xuezhiw, yaoqin, jinlinc\}@google.com \quad kwchang@cs.ucla.edu
}

\begin{document}
\maketitle
\begin{abstract}
Large pre-trained language models have shown remarkable performance over the past few years. These models, however, sometimes learn  superficial features from the dataset and cannot generalize to the distributions that are dissimilar to the training scenario. There have been several approaches proposed to reduce model's reliance on these bias features which can improve  model robustness in the out-of-distribution setting. However, existing methods usually use a fixed low-capacity model to deal with various bias features, which ignore the learnability of those features. In this paper, we analyze a set of existing bias features and demonstrate there is no single model that works best for all the cases. We further show that by choosing an appropriate bias model, we can obtain a better  robustness result than  baselines with a more sophisticated model design.
\end{abstract}

\section{Introduction}
Advances in pre-trained language models have shown great performance in natural language processing (NLP) benchmarks.
However, these models often learn dataset-specific patterns and cannot generalize well to out-of-distribution data~\cite{mccoy2019right,niven2019probing,si2019does,ko2020look}. These patterns are referred to as \textit{bias features}, which have strong indications of instance labels but do not necessarily generalize to out-of-distribution data \cite{Geirhos_2020}. For example, in MNLI~\cite{williams2018broad}, the appearance of a negation word in an example has a strong correlation with label ``contradiction''~\cite{gururangan2018annotation}.  A model leveraging such bias features can exhibit  good performance on in-domain data but will break when evaluated on an out-of-distribution test set where the correlation between the patterns and labels no longer holds.

Given prior knowledge of possible bias features in the dataset, several approaches have been proposed to reduce  models' overreliance on the bias features~\cite{clark2019dont, he2019unlearn,utama2020mind}. 
The underlying idea  is to discourage the model to learn from ``easy'' examples that can be  predicted correctly solely based on bias features. These works first train a \textit{bias model} to capture bias patterns. They then train a \textit{main model} and ensemble it with the bias model in a way that the predictions of the main model are adjusted by not leveraging the strategy captured by the bias model. 
\textit{Product-of-experts}~\cite{hinton2002training} and \textit{self-distillation} approaches have been widely adopted for the ensemble~\cite{clark2019dont, he2019unlearn, mahabadi2019simple, utama2020mind,du2021towards}.

Although these methods improve model robustness on some benchmark datasets, none of them study how to choose the bias model and only assume that a weak classifier (e.g., logistic regression) can explicitly capture the bias patterns. We argue that it is very important to choose the appropriate bias model. On one hand,  different bias features may not be captured by one model with a fixed model size (capacity). For example, in MNLI, a model capturing the ``negation word occurrence''  does not necessarily capture the token overlap pattern at the same time. On the other hand, we do not expect an over-capacity bias model as it may capture non-bias features which will also be factored out during the ensemble, thus worsening the overall model performance. To do this, we train  bias models with different capacities leveraging the \textit{product-of-experts} method and compare with current state-of-the-art methods.
We empirically show the different learnability of the bias patterns requires models with diverse capacities to better capture them. For example, a BERT-mini model can learn a token-overlap style bias pattern better in the MNLI dataset compared with the   logistic regression used by  existing literature. 

In this work, we conduct a deep analysis of the existing literature on ensemble-based methods for model robustness improvement where most of them follow the hypothesis of adopting one simple model to learn bias patterns. Instead, we study different bias models and demonstrate their ability to capture the bias patterns varies.
We propose an approach to selecting the ``best'' bias model by splitting the development set into ``easy'' and ``challenge'' subsets. By utilizing the best bias models for different bias features, we show that we can make the models more robust compared to existing baselines, on both natural language inference and fact-verification tasks.

\section{Bias Model Selection}
To understand to what extent our models can conquer the biases, for each bias, we create a bias training dataset only consisting of examples with this bias feature to reduce the impact of other biases. To evaluate model's ability to overcome the bias, we create a corresponding  challenge test set which is composed of examples with that specific bias feature and not following the distribution in the training set. Examples of such  sets are in  Sec.~\ref{subsec:bias_challenge_set}.

\begin{figure}[!t]
 \includegraphics[width=0.95\columnwidth]{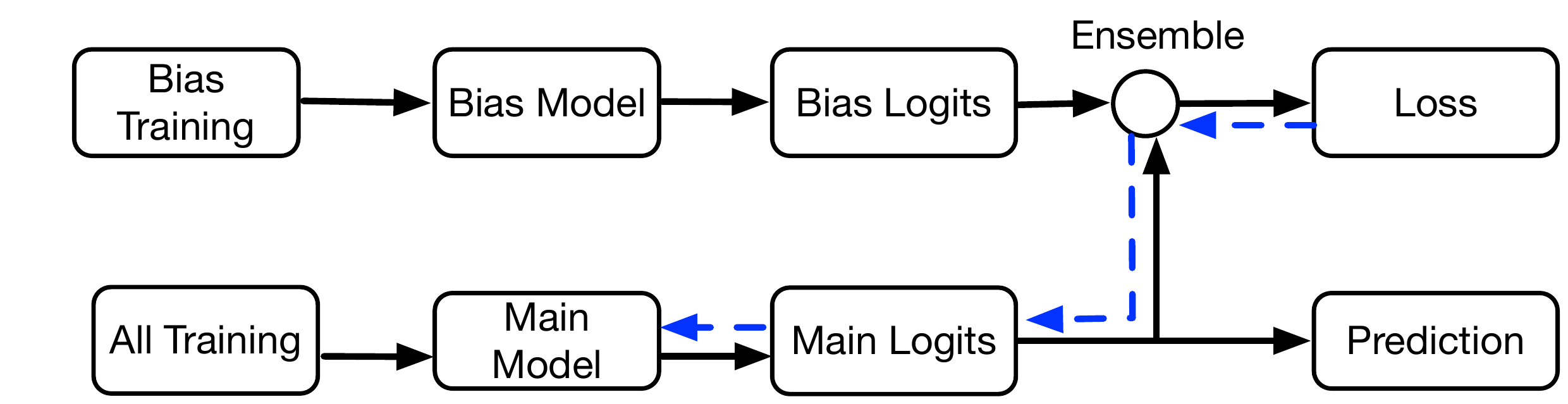}
 \caption{An overview of model pipeline. Bias model is trained on biased subset of the original training dataset (more details are in Sec.~\ref{subsec:bias_challenge_set}). Blue arrows stand for the gradient flow. We only use the main model when doing the evaluation.}
 \label{fig:model}
\end{figure}

To obtain the best bias  model for each bias feature, we first train various bias  models with different capacities on the  bias training  set and ensemble them with the main  model. We then evaluate the main model on the corresponding challenge set and choose the one exhibiting the best robustness result. Our model pipeline is shown in Figure~\ref{fig:model}. 

\setlength{\abovedisplayskip}{5pt}
\setlength{\belowdisplayskip}{5pt}
In this work, following~\citet{clark2019dont}, we leverage \textit{product-of-experts} (PoE), a commonly used method for model robustness improvement, as an example to illustrate our argument.\footnote{Various methods can be chosen for the analysis such as \textit{learned-mixin} which is the adaptive version of PoE. However, as stated in \citet{clark2019dont}, the \textit{learned-mixin} method could be unstable in some cases.} 
Given a dataset $\mathcal{D}=\{(x_i, y_i)\}_{i\in[1,n]}$, where $y_i\in \{1, 2, \ldots, C\}$, a bias model  $h(x_i;\theta_b) = \langle b_{i1}, b_{i2}, \ldots, b_{iC} \rangle$ and a main model $f(x_i; \theta)=\langle p_{i1}, p_{i2}, \ldots, p_{iC}\rangle$ where $b_{ij}$ and $p_{ij}$ are the probabilities predicted for label category $j$, the goal is to learn $\theta$ that can make a correct prediction for an input example without using the patterns learned by the bias model. To achieve such a goal, PoE~\cite{hinton2015distilling} fuses the two models as $$\hat{p_i} = \text{softmax}(\log(p_i) + \log(b_i)).$$
During training, the model is optimized for the cross-entropy loss based on $\hat{p}$. After training, only the main model  $f$ will be used for evaluation.

\section{Experiment}

In this section, we use two tasks (in English) to study the effects of different bias models. We analyze the best bias model for various bias features. 

\subsection{Dataset}
\paragraph{Fact Verification.}
FEVER~\cite{thorne2018fever} is a dataset for fact verification task. Each instance contains a claim sentence and an evidence sentence. The goal is to verify if the claim is Supported, Refuted or NotEnoughInfo with the evidence. We evaluate  model robustness on  Fever-Symmetric dataset~\cite{schuster2019towards}.

\paragraph{Natural Language Inference.} The goal of the natural language inference (NLI) task is to identify the relationship (Entailment, Contradiction or Neutral) between the hypothesis and premise sentences. 
We use the MNLI dataset~\cite{williams2018broad} for training and evaluate the model robustness on the HANS dataset~\cite{mccoy2019right}.

\subsection{Bias Features}
\newcite{schuster2019towards} show that for the FEVER dataset, only using the claim sentence can obtain comparable results to using both claim and evidence. Hence we use this \textsc{claim-only} feature as one of our bias features. Similarly, we add another type of bias feature, which only uses the evidence sentence to make the prediction, and we refer to this as \textsc{evidence-only} bias.

\citet{clark2019dont} list several bias features in MNLI dataset, such as whether all the tokens of the hypothesis appear in the premise (\textsc{all-in-p}), whether the hypothesis is a subsequence in premise (\textsc{h-is-subseq}), the percentage of words in hypothesis that are also in premise (\textsc{percent-in-p}), and some bias features based on word embeddings. In this work, we study the first two  and in addition, we also consider a ``\textsc{neg-in-h}'' bias which refers to having negation words  in the hypothesis~\cite{gururangan2018annotation}.

\begin{table}[!t]
\centering
\begin{tabular}{lcc}
\toprule
Bias Model & FEVER\textsubscript{Dev} & \textsc{claim-only}  \\
\hline
None &$86.10_{\pm 0.13}$ & $82.53_{\pm 0.48} $\\
MLP\textsubscript{claim} &$90.25_{\pm0.44}$ &  $87.78_{\pm 0.24}$ \\
BERT\textsubscript{tiny} & $86.85_{\pm0.50}$& $86.89_{\pm 0.79}$ \\
BERT\textsubscript{mini} & $83.56_{\pm0.58 }$& $86.81_{\pm 0.55}$ \\
BERT\textsubscript{small} &$84.52_{\pm 0.20} $&  $87.67_{\pm 0.12}$ \\
BERT\textsubscript{medium} & $83.78_{\pm 0.38}$&  $87.15_{\pm 0.73}$ \\
BERT\textsubscript{base} & $86.15_{\pm 0.59}$& ${\bf 89.82}_{\pm 0.89}$ \\
\bottomrule
\end{tabular}
\caption{Model evaluation results with different bias models for \textsc{claim-only} bias in FEVER. The values are accuracy scores (average\textsubscript{ $\pm$ standard deviation}, in \%) over 3 runs on FEVER dev and \textsc{claim-only} challenge set. The best bias model for the \textsc{claim-only} bias is a BERT-base model. }
\label{tab:fever_claimonly}
\end{table}

\begin{table}[!t]
\centering
\begin{tabular}{lcc}
\toprule
Bias Model &FEVER\textsubscript{Dev} & \textsc{evidence-only}  \\
\hline
None &$86.10_{\pm 0.13}$ & $88.10_{\pm 0.79}$ \\
MLP\textsubscript{evidence} &$92.47_{\pm0.08}$ &  $94.18_{\pm 1.21}$ \\
BERT\textsubscript{tiny} & $93.37_{\pm 0.31}$& ${\bf 96.03}_{\pm 1.37}$ \\
BERT\textsubscript{mini} & $93.13_{\pm 0.24 }$& $94.97_{\pm 1.21}$ \\
BERT\textsubscript{small} &$92.74_{\pm 0.06} $&  $94.44_{\pm 0.79}$ \\
BERT\textsubscript{medium} & $93.12_{\pm 0.10}$&  $94.44_{\pm 2.10}$ \\
BERT\textsubscript{base} & $92.02_{\pm 0.22}$& $93.92_{\pm 0.92}$ \\
\bottomrule
\end{tabular}
\caption{Model evaluation results with different bias models for \textsc{evidence-only} in FEVER. The values are the averaged accuracy scores (in \%) on FEVER Dev and \textsc{evidence-only} challenge set over 3 runs. To deal with \textsc{evidence-only} bias feature, the best bias model is a BERT-tiny model.}
\label{tab:fever_evidenceonly}
\end{table}

\subsection{Bias and Challenge Sets}
\label{subsec:bias_challenge_set}

Our method leverages the known bias features and verifies model's ability to overcome the biases based on the bias training and challenge test sets.
 To obtain the biased training set for MNLI, we follow \citet{clark2019dont} to select the examples equipped for each bias features  separately. For both \textsc{all-in-p} and \textsc{h-is-subseq}, they are strongly related to ``entailment'' label while \textsc{neg-in-h} is closely related to ``contradiction'' label.  To build the \textsc{challenge set}, we filter test examples with a corresponding bias (e.g., satisfying the \textsc{all-in-p}) but the label does not follow the bias pattern in the training dataset (e.g., labels are not ``entailment''). 
 In FEVER, there is no straightforward way to determine if one example can be purely predicted by one sentence. To create the bias training set for the \textsc{claim-only}, we first fine-tune a BERT-base model on the FEVER dataset. We then make the  prediction based only on the claim sentence and collect all the examples that can be predicted correctly as the biased training set. To build the challenge set, we collect examples that cannot be correctly predicted by only looking at the claim sentence from the dev set. We do the same for the \textsc{evidence-only} bias in  FEVER. 
 For these two tasks, the bias training set is obtained from the corresponding task training set and the challenge set is obtained from the task dev (or test) set. Our bias only models are trained on the bias training dataset and evaluated on the bias challenge set.

\begin{table}[!t]
\centering
\begin{adjustbox}{width=\columnwidth}
\begin{tabular}{lccc} 
\toprule
Method  & FEVER\textsubscript{Dev} &Symm\textsubscript{v1} & Symm\textsubscript{v2}  \\ 
\midrule
 BERT-base  &  $86.10_{\pm 0.13}$  &  $58.34_{\pm 2.22}$   &     $65.26_{\pm 0.77}$   \\
 \midrule
PoE\textsubscript{MLP} & $86.26_{\pm 0.09}$ & $59.93_{\pm 0.81}$ & $64.93_{\pm 1.27}$ \\
SelfDistill & $85.13_{\pm 0.40}$ & $55.65_{\pm 0.56}$ & $62.55_{\pm 0.53}$ \\
Reweight  & $85.20_{\pm 0.38}$ & $58.86_{\pm 1.09}$ & $64.79_{\pm 0.57}$ \\
PoE\textsubscript{Ours} & $\bf{93.33}_{\pm 0.30}$ & $\bf{69.08}_{\pm 2.02} $ & $\bf{73.46}_{\pm 1.70}$ \\
\bottomrule
\end{tabular}
\end{adjustbox}
\caption{Robustness results on FEVER when fusing all the bias features. {BERT-base} means the baseline model without fusing a bias model.}
\label{tab:fever_fuse}
\end{table}

 \subsection{Capacity for Bias Models}
In this section, we verify the best bias model for each bias feature. We use a BERT-base model as the main model for both the FEVER and MNLI datasets. In terms of the capacity for the bias model, we consider different BERT models (from tiny to base) as well as the one used in the existing literature. For MNLI, a widely used bias model is a logistic regression model trained on some predefined biased features ~\cite{clark2019dont}.  For FEVER dataset, existing work~\cite{utama2020mind} leverages a shallow non-linear classifier, in this work, we use a multilayer perceptron (MLP) model.

Table~\ref{tab:fever_claimonly} shows the results on FEVER when we use the bias models with different capacities, from   MLP to  BERT-base. The first row ``None'' stands for a naive BERT-base model without any bias  model. 
\textsc{claim-only} stands for the model's performance on the challenge set we create. It suggests that to deal with the \textsc{claim-only} bias, using BERT-base  as the bias model can better improve the robustness on the challenge set, at the same time, we see the model keeps its performance on the in-distribution dev set.
Similarly, when dealing with the \textsc{evidence-only} bias, in Table~\ref{tab:fever_evidenceonly} we see that using BERT-tiny as the bias model can improve the model robustness better than other choices without performance loss in the in-distribution data.

\begin{table}[!t]
\centering
\begin{adjustbox}{width=\columnwidth}
\begin{tabular}{lccc}
\toprule
Bias Model & dev-m & dev-mm & \textsc{all-in-p}    \\
\hline
None &$83.78_{\pm 0.24}$ & $84.21_{\pm 0.11} $ & $28.60_{\pm 5.51}$  \\
Logistic Reg. &$83.52_{\pm 0.13}$ &  $83.80_{\pm 0.12}$ & $38.87_{\pm 1.77}$   \\
BERT\textsubscript{tiny} & $80.02_{\pm 0.83}$& $80.87_{\pm 0.30}$ & $ {\bf 45.27}_{\pm 3.19}$  \\
BERT\textsubscript{mini} & $80.72_{\pm 0.09}$& $80.78_{\pm 0.37}$ & $43.03_{\pm 3.78}$  \\
BERT\textsubscript{small} &$81.98_{\pm 0.42} $&  $82.27_{\pm 0.14}$ & $40.67_{\pm 1.70}$  \\
BERT\textsubscript{medium} & $81.97_{\pm 0.34}$&  $82.40_{\pm 0.14}$ & $36.80_{\pm 0.85}$  \\
BERT\textsubscript{base} & $82.71_{\pm 0.35}$& ${ 83.17}_{\pm 0.32}$ & $37.77_{\pm 5.2}$ \\
\bottomrule
\end{tabular}
\end{adjustbox}
\caption{Model accuracy on MNLI dev-matched, dev-mismatched  and corresponding challenge set when dealing with \textsc{all-in-p} bias. BERT-tiny is the best bias model for this bias feature. More results about other bias features are in appendix.}
\label{tab:mnli_biasonly}
\end{table}

We compare our methods with two other baselines, one is \texttt{Reweight}, where a model is trained on the weighted dataset. The weight of example $x_i$ is $1 - b_{i{y_i}}$ where $b_{i{y_i}}$ is the probability from the bias model on the correct label $y_i$~\cite{clark2019dont}. Another baseline is  \texttt{self-distillation}~\cite{utama2020mind}, where the ensemble is based on the knowledge distillation~\cite{hinton2015distilling}. It is a more complicated ensemble method than PoE and requires an additional teacher model. More details are provided in the Appendix.

We show that by fusing different bias logits together, we can provide a way to improve the model robustness. We compare our methods with existing literature which leverages the biases from a fixed model~\cite{utama2020mind}.  In Table~\ref{tab:fever_fuse}, PoE\textsubscript{MLP} refers to a baseline which uses the MLP as the bias model.\footnote{For the self-distillation, we use the logits released by \citet{utama2020mind}}   PoE\textsubscript{Ours} fuses a BERT-base model with weighted bias logits from our claim-only and evidence-only bias models and it outperforms the baselines by over 8\% on the test benchmarks.

For MNLI, we consider the same set of BERT models. All of the bias models are trained on the examples that have this bias feature. 
The results are shown in Table~\ref{tab:mnli_biasonly}, where we demonstrate again that the best bias model dealing with one bias feature (e.g., \textsc{all-in-p}) does not necessarily work best for another one (e.g. \textsc{neg-in-h}). By mixing the logits from the best bias models for each bias feature, a PoE method can outperform  self-distillation which has a more sophisticated ensemble schema.

\paragraph{Insights} We notice that the logits fused into the main models can play a significant role in the model performance. For example, we observe that an overconfident bias model can hurt the model performance in the in-distribution evaluation set. 
We also find a possible negative effect on the robustness result when dealing with the bias features not related to the test set. For example, in HANS, there are no instances with the \textsc{neg-in-h} bias, and fusing the logits from the bias model for \textsc{neg-in-h} sometimes hurts the performance on HANS (more in appendix). However, in most cases, we do not have access to the bias features in the test set, how to deal with the potential conflicts between different bias features remains an unexplored yet very important direction and we leave it for future study.

\begin{table}[!t]
\begin{adjustbox}{width=\columnwidth}
\begin{tabular}{lccc} 
\toprule
Method  & dev-m  &dev-mm & HANS  \\ 
\midrule
 BERT-base  &  $83.78_{\pm 0.24}$  &  $84.21_{\pm 0.11}$   &     $63.05_{\pm 3.07}$   \\
 \midrule
 PoE\textsubscript{LogisticReg.} & $83.52_{\pm 0.13}$ & $83.80_{\pm 0.12}$ & $67.07_{\pm 1.27}$ \\
SelfDistill & $84.74_{\pm 0.27}$ & $85.19_{\pm 0.16}$ & $70.51_{\pm 0.63}$ \\
Reweight  & $83.89_{\pm 0.09}$ & $84.06_{\pm 0.30}$ & $ 65.10_{\pm  2.75}$ \\
PoE\textsubscript{Ours} & $ {81.46}_{\pm 0.38}$ & $ {81.63}_{\pm 0.17} $ & $\bf{70.58}_{\pm 1.10}$ \\
\bottomrule
\end{tabular}
\end{adjustbox}
\caption{Robustness results for combining all bias features in MNLI. {BERT-base} means the baseline model without fusing a bias model.}
\label{table:mnli_fuse}
\end{table}

\section{Related Work}
Recent NLP models show great performance when evaluating on the in-distribution test set. However, such results might not hold when the test set is out-of-distribution. 
For example, \citet{schuster2019towards} discover that a fact verification model may make a prediction by looking at the occurrence of certain phrases in the input example. Similar scenarios have been observed in other applications such as in visual question answering~\cite{agrawal2018don},  and paraphrase identification~\cite{zhang2019paws}.

Several ensemble-based methods have shown improvement in model robustness when dealing with dataset biases~\cite{clark2019dont, he2019unlearn,mahabadi2019simple,utama2020mind}. Such methods usually contain two components for the ensemble and focus on different ensemble strategies. In contrast, we analyze the components for  the ensemble. Our work fills in the gap of a deeper understanding of the bias patterns in the dataset  and  provide a pipeline to choose the best component for the ensemble so that we can improve model robustness.

\section{Discussion}
How to improve the model robustness has been an important research topic, and several approaches have been proposed for such a goal, ranging from dataset augmentation~\cite{kaushik2019learning} to model architecture design~\cite{lewis2018generative}. Although several ensemble-based methods have shown great improvement, they treat all the dataset artifacts exactly the same way. In this work, we revisit such methods,  and demonstrate that, not all the dataset artifacts are the same and they require different capacity models to deal with. 
Contrary to the common beliefs in the existing literature that a smaller-capacity model captures the bias features, our paper is the first that investigates the effect of the bias model size and shows that better robustness needs to be achieved by bias models with different capacities. 
We also show that by better leveraging the information learned from the dataset artifacts, a simple ensemble method can achieve a better or the same level of model robustness.

\section*{Limitations}
Our study in this paper only considers some known bias features. While this setting is common in the literature, we argue that in real applications, it might be very hard to get such information. 
In addition, this work is based on the ensemble method for robustness improvement and the bias models obtained for PoE might not be the same for another method. 
There are other ways to achieve robustness such as adversarial training~\cite{grand2019adversarial} which we leave for future study to show how they can be used to deal with various bias features.


\section*{Acknowledgements}
We would like to thank all the reviewers for their valuable feedback. 

\bibliography{nlp,preprint,ref,custom}

\begin{thebibliography}{22}
\expandafter\ifx\csname natexlab\endcsname\relax\def\natexlab#1{#1}\fi

\bibitem[{Agrawal et~al.(2018)Agrawal, Batra, Parikh, and
  Kembhavi}]{agrawal2018don}
Aishwarya Agrawal, Dhruv Batra, Devi Parikh, and Aniruddha Kembhavi. 2018.
\newblock Don't just assume; look and answer: Overcoming priors for visual
  question answering.
\newblock In \emph{2018 IEEE/CVF Conference on Computer Vision and Pattern
  Recognition}, pages 4971--4980. IEEE.

\bibitem[{Clark et~al.(2019)Clark, Yatskar, and Zettlemoyer}]{clark2019dont}
Christopher Clark, Mark Yatskar, and Luke Zettlemoyer. 2019.
\newblock Don{'}t take the easy way out: Ensemble based methods for avoiding
  known dataset biases.
\newblock In \emph{Proceedings of the 2019 Conference on Empirical Methods in
  Natural Language Processing and the 9th International Joint Conference on
  Natural Language Processing (EMNLP-IJCNLP)}, pages 4069--4082.

\bibitem[{Du et~al.(2021)Du, Manjunatha, Jain, Deshpande, Dernoncourt, Gu, Sun,
  and Hu}]{du2021towards}
Mengnan Du, Varun Manjunatha, Rajiv Jain, Ruchi Deshpande, Franck Dernoncourt,
  Jiuxiang Gu, Tong Sun, and Xia Hu. 2021.
\newblock Towards interpreting and mitigating shortcut learning behavior of
  {NLU} models.
\newblock In \emph{Proceedings of the 2021 Conference of the North American
  Chapter of the Association for Computational Linguistics: Human Language
  Technologies}, pages 915--929.

\bibitem[{Geirhos et~al.(2020)Geirhos, Jacobsen, Michaelis, Zemel, Brendel,
  Bethge, and Wichmann}]{Geirhos_2020}
Robert Geirhos, Jörn-Henrik Jacobsen, Claudio Michaelis, Richard Zemel,
  Wieland Brendel, Matthias Bethge, and Felix~A. Wichmann. 2020.
\newblock \href {https://doi.org/10.1038/s42256-020-00257-z} {Shortcut learning
  in deep neural networks}.
\newblock \emph{Nature Machine Intelligence}, 2(11):665–673.

\bibitem[{Grand and Belinkov(2019)}]{grand2019adversarial}
Gabriel Grand and Yonatan Belinkov. 2019.
\newblock Adversarial regularization for visual question answering: Strengths,
  shortcomings, and side effects.
\newblock In \emph{Proceedings of the Second Workshop on Shortcomings in Vision
  and Language}, pages 1--13.

\bibitem[{Gururangan et~al.(2018)Gururangan, Swayamdipta, Levy, Schwartz,
  Bowman, and Smith}]{gururangan2018annotation}
Suchin Gururangan, Swabha Swayamdipta, Omer Levy, Roy Schwartz, Samuel Bowman,
  and Noah~A. Smith. 2018.
\newblock Annotation artifacts in natural language inference data.
\newblock In \emph{Proceedings of the 2018 Conference of the North {A}merican
  Chapter of the Association for Computational Linguistics: Human Language
  Technologies, Volume 2 (Short Papers)}, pages 107--112.

\bibitem[{He et~al.(2019)He, Zha, and Wang}]{he2019unlearn}
He~He, Sheng Zha, and Haohan Wang. 2019.
\newblock Unlearn dataset bias in natural language inference by fitting the
  residual.
\newblock In \emph{Proceedings of the 2nd Workshop on Deep Learning Approaches
  for Low-Resource NLP (DeepLo 2019)}, pages 132--142.

\bibitem[{Hinton et~al.(2015)Hinton, Vinyals, Dean
  et~al.}]{hinton2015distilling}
Geoffrey Hinton, Oriol Vinyals, Jeff Dean, et~al. 2015.
\newblock Distilling the knowledge in a neural network.
\newblock \emph{arXiv preprint arXiv:1503.02531}, 2(7).

\bibitem[{Hinton(2002)}]{hinton2002training}
Geoffrey~E Hinton. 2002.
\newblock Training products of experts by minimizing contrastive divergence.
\newblock \emph{Neural computation}, 14(8):1771--1800.

\bibitem[{Kaushik et~al.(2019)Kaushik, Hovy, and Lipton}]{kaushik2019learning}
Divyansh Kaushik, Eduard Hovy, and Zachary Lipton. 2019.
\newblock Learning the difference that makes a difference with
  counterfactually-augmented data.
\newblock In \emph{International Conference on Learning Representations}.

\bibitem[{Ko et~al.(2020)Ko, Lee, Kim, Kim, and Kang}]{ko2020look}
Miyoung Ko, Jinhyuk Lee, Hyunjae Kim, Gangwoo Kim, and Jaewoo Kang. 2020.
\newblock Look at the first sentence: Position bias in question answering.
\newblock In \emph{Proceedings of the 2020 Conference on Empirical Methods in
  Natural Language Processing (EMNLP)}, pages 1109--1121.

\bibitem[{Lewis and Fan(2018)}]{lewis2018generative}
Mike Lewis and Angela Fan. 2018.
\newblock Generative question answering: Learning to answer the whole question.
\newblock In \emph{International Conference on Learning Representations}.

\bibitem[{Mahabadi and Henderson(2019)}]{mahabadi2019simple}
Rabeeh~Karimi Mahabadi and James Henderson. 2019.
\newblock Simple but effective techniques to reduce biases.

\bibitem[{McCoy et~al.(2019)McCoy, Pavlick, and Linzen}]{mccoy2019right}
Tom McCoy, Ellie Pavlick, and Tal Linzen. 2019.
\newblock Right for the wrong reasons: Diagnosing syntactic heuristics in
  natural language inference.
\newblock In \emph{Proceedings of the 57th Annual Meeting of the Association
  for Computational Linguistics}, pages 3428--3448.

\bibitem[{Niven and Kao(2019)}]{niven2019probing}
Timothy Niven and Hung-Yu Kao. 2019.
\newblock Probing neural network comprehension of natural language arguments.
\newblock In \emph{Proceedings of the 57th Annual Meeting of the Association
  for Computational Linguistics}, pages 4658--4664.

\bibitem[{Schuster et~al.(2019)Schuster, Shah, Yeo, Roberto Filizzola~Ortiz,
  Santus, and Barzilay}]{schuster2019towards}
Tal Schuster, Darsh Shah, Yun Jie~Serene Yeo, Daniel Roberto Filizzola~Ortiz,
  Enrico Santus, and Regina Barzilay. 2019.
\newblock Towards debiasing fact verification models.
\newblock In \emph{Proceedings of the 2019 Conference on Empirical Methods in
  Natural Language Processing and the 9th International Joint Conference on
  Natural Language Processing (EMNLP-IJCNLP)}, pages 3419--3425.

\bibitem[{Si et~al.(2019)Si, Wang, Kan, and Jiang}]{si2019does}
Chenglei Si, Shuohang Wang, Min-Yen Kan, and Jing Jiang. 2019.
\newblock What does bert learn from multiple-choice reading comprehension
  datasets?
\newblock \emph{arXiv preprint arXiv:1910.12391}.

\bibitem[{Thorne et~al.(2018)Thorne, Vlachos, Christodoulopoulos, and
  Mittal}]{thorne2018fever}
James Thorne, Andreas Vlachos, Christos Christodoulopoulos, and Arpit Mittal.
  2018.
\newblock {FEVER}: a large-scale dataset for fact extraction and
  {VER}ification.
\newblock In \emph{Proceedings of the 2018 Conference of the North {A}merican
  Chapter of the Association for Computational Linguistics: Human Language
  Technologies, Volume 1 (Long Papers)}, pages 809--819.

\bibitem[{Utama et~al.(2020)Utama, Moosavi, and Gurevych}]{utama2020mind}
Prasetya~Ajie Utama, Nafise~Sadat Moosavi, and Iryna Gurevych. 2020.
\newblock Mind the trade-off: Debiasing {NLU} models without degrading the
  in-distribution performance.
\newblock In \emph{Proceedings of the 58th Annual Meeting of the Association
  for Computational Linguistics}, pages 8717--8729.

\bibitem[{Williams et~al.(2018)Williams, Nangia, and
  Bowman}]{williams2018broad}
Adina Williams, Nikita Nangia, and Samuel Bowman. 2018.
\newblock A broad-coverage challenge corpus for sentence understanding through
  inference.
\newblock In \emph{Proceedings of the 2018 Conference of the North American
  Chapter of the Association for Computational Linguistics: Human Language
  Technologies, Volume 1 (Long Papers)}, pages 1112--1122.

\bibitem[{Wolf et~al.(2019)Wolf, Debut, Sanh, Chaumond, Delangue, Moi, Cistac,
  Rault, Louf, Funtowicz et~al.}]{wolf2019huggingface}
Thomas Wolf, Lysandre Debut, Victor Sanh, Julien Chaumond, Clement Delangue,
  Anthony Moi, Pierric Cistac, Tim Rault, R{\'e}mi Louf, Morgan Funtowicz,
  et~al. 2019.
\newblock Huggingface's transformers: State-of-the-art natural language
  processing.
\newblock \emph{arXiv preprint arXiv:1910.03771}.

\bibitem[{Zhang et~al.(2019)Zhang, Baldridge, and He}]{zhang2019paws}
Yuan Zhang, Jason Baldridge, and Luheng He. 2019.
\newblock {PAWS}: Paraphrase adversaries from word scrambling.
\newblock In \emph{Proceedings of the 2019 Conference of the North {A}merican
  Chapter of the Association for Computational Linguistics: Human Language
  Technologies, Volume 1 (Long and Short Papers)}, pages 1298--1308.

\end{thebibliography}
\bibliographystyle{acl_natbib}

\appendix

\section{Appendix}
\label{sec:appendix}
\paragraph{Training details} All the BERT model we used is from HuggingFace~\cite{wolf2019huggingface}. We use the default hyperparameters in \citet{utama2020mind} to train the FEVER model, e.g., 3 epochs with a learning rate $5 \times 10^{-5}$. To train the FEVER model, we leverage the preprocessed training data as indicated in~\citet{schuster2019towards}. For the MNLI dataset, we train the model using the default hypterparameters except that we train the model for 6 epochs to make all the models converge. All the results in the paper are the averaged value for 3 runs.
We train our models with NVIDIA GeForce RTX 2080 Ti GPUs on Pytorch, and each epoch takes approximate 1 hour. 

\paragraph{Self-Distillation} Given a teacher model, which provides the prediction for input example $x_i$ as $\langle \hat{p}_{i1}, \dots, \hat{p}_{iC} \rangle $, with the probability from the bias model $\langle b_{i1}, \dots, b_{iC} \rangle$, the scaled teacher output for each example is computed as $\mathbf{s} = \langle s_{i1}, \dots, s_{iC} \rangle$, where

$$s_{ij} = \frac{\hat{p}_{ij}^{(1-b_{iy_i})}}{\sum_{k=1}^{C} \hat{p}_{ik}^{(1-b_{iy_i})}}.$$

\noindent Then a model is trained to minimize the cross entropy loss between the scaled teacher output and the current output of the main model.

\paragraph{Bias model results} We list the results on the dev and corresponding \textsc{challenge} sets for the MNLI task when training on the \textsc{h-is-subseq} (in Table~\ref{tab:dev_mnli_hissubseq}) and \textsc{neg-in-h} (in Table~\ref{tab:dev_mnli_neginh}) bias features separately.

\begin{table}[!h]
\centering
\begin{adjustbox}{width=\columnwidth}
\begin{tabular}{lccc}
\toprule
Bias Model &dev-m & dev-mm & \textsc{h-is-subseq} \\
\hline
None &$86.10_{\pm 0.13}$ & $88.10_{\pm 0.79}$ & $ 14.40_{\pm 4.44 }$\\
Logistic Reg. &$92.47_{\pm0.08}$ &  $94.18_{\pm 1.21}$  & $26.47_{\pm 2.86}$\\
BERT\textsubscript{tiny} & $81.45_{\pm 0.27}$& ${ 81.75}_{\pm 0.55}$ & $29.53_{\pm 3.87}$\\
BERT\textsubscript{mini} & $79.51_{\pm 0.10}$& $80.35_{\pm 0.11}$ & ${\bf 33.77}_{\pm 4.17}$ \\
BERT\textsubscript{small} &$80.29_{\pm 0.10} $&  $80.84_{\pm 0.11}$ & $30.70_{\pm 7.64}$\\
BERT\textsubscript{medium} & $80.39_{\pm 0.24}$&  $80.94_{\pm 0.24}$ & $27.13_{\pm 1.35}$ \\
BERT\textsubscript{base} & $81.47_{\pm 0.27}$& $81.69_{\pm 0.39}$ & $29.93_{\pm 5.20}$ \\
\bottomrule
\end{tabular}
\end{adjustbox}
\caption{Results on dev and challenge sets for MNLI dataset when dealing with the \textsc{h-is-subseq} bias feature.}
\label{tab:dev_mnli_hissubseq}
\end{table}

\begin{table}[!h]
\centering
\begin{adjustbox}{width=\columnwidth}
\begin{tabular}{lccc}

\toprule
Bias Model &dev-m & dev-mm & \textsc{neg-in-h} \\
\hline
None &$86.10_{\pm 0.13}$ & $88.10_{\pm 0.79}$ & $ 77.15_{\pm 0.47 }$\\
Logistic Reg. &$92.47_{\pm0.08}$ &  $94.18_{\pm 1.21}$ & ${\bf 77.69}_{\pm 1.01}$\\
BERT\textsubscript{tiny} & $82.87_{\pm 0.21}$& ${ 83.29}_{\pm 0.05}$ & $77.06_{\pm 0.77}$\\
BERT\textsubscript{mini} & $82.63_{\pm 0.15}$& $82.91_{\pm 0.25}$ & $75.18_{\pm 1.10}$\\
BERT\textsubscript{small} &$82.00_{\pm 0.10} $&  $82.13_{\pm 0.06}$ &$73.29_{\pm 0.16}$ \\
BERT\textsubscript{medium} & $80.33_{\pm 0.37}$&  $80.57_{\pm 0.27}$& $72.37_{\pm 1.40}$\\
BERT\textsubscript{base} & $83.55_{\pm 0.12}$& $84.06_{\pm 0.23}$& $60.84_{\pm 2.89}$ \\
\bottomrule
\end{tabular}
\end{adjustbox}
\caption{Results on dev and challenge sets for MNLI dataset when dealing with the \textsc{neg-in-h} bias feature.}
\label{tab:dev_mnli_neginh}
\end{table}

\paragraph{Bias features vs. Model robustness} We show here that dealing with the \textsc{neg-in-h} bias can potentially have negative effect on the test result. For example, when we fuse the logits from a BERT-tiny model only on the examples having the \textsc{neg-in-h} bias feature, we observe the accuracy on HANS drops from $63.05\%$ to $59.93\%$, while fusing \textsc{all-in-p} or \textsc{h-is-subseq} achieves the accuracy $65.39\%$ and $63.65\%$ respectively.

\end{document}